\documentclass[conference]{IEEEtran}
\IEEEoverridecommandlockouts
\usepackage{xcolor}
\usepackage{cite}
\usepackage{caption}  
\usepackage{amsmath,amssymb,amsfonts}
\usepackage{algorithmic}
\usepackage{graphicx}
\usepackage{hyperref}
\usepackage{textcomp}
\usepackage{mathrsfs}
\usepackage{booktabs} 
\usepackage{tikz}
\usepackage{pgfplots}
\usepackage{cleveref}
\usepackage{dblfloatfix}
\pgfplotsset{compat=1.18}
\usepackage{float}
\usepackage{xcolor}
\def\BibTeX{{\rm B\kern-.05em{\sc i\kern-.025em b}\kern-.08em
    T\kern-.1667em\lower.7ex\hbox{E}\kern-.125emX}}

\definecolor{darkgreen}{rgb}{0, 0.5, 0.1}

\begin{document}

\title{Generalized Category Discovery in Hyperspectral Images via Prototype Subspace Modeling\\

\thanks{This work was supported in part by the Bijzonder Onderzoeksfonds (BOF) under Grant BOF.24Y.2021.0049.01, in part by the Flanders AI Research Programme under Grant 174B09119, and in part by the China Scholarship Council under Grant 202106150007.}
}

\author{
    \IEEEauthorblockN{
        Xianlu Li\IEEEauthorrefmark{1}, 
        Nicolas Nadisic\IEEEauthorrefmark{1}\IEEEauthorrefmark{3}, Shaoguang Huang\IEEEauthorrefmark{2}, and
        Aleksandra  Pi\v{z}urica\IEEEauthorrefmark{1}
    }
    \IEEEauthorblockA{
        \IEEEauthorrefmark{1}Department of Telecommunications and Information Processing, Ghent University, Belgium.\\
    }
    \IEEEauthorblockA{
        \IEEEauthorrefmark{3}Royal Institute for Cultural Heritage (KIK-IRPA), Brussels, Belgium.\\
    }
    \IEEEauthorblockA{
        \IEEEauthorrefmark{2}School of Computer Science, China University of Geosciences, Wuhan, China.\\
    }
}


\maketitle

\begin{abstract}
Generalized category discovery~(GCD) seeks to jointly identify both known and novel categories in unlabeled data. While prior works have mainly focused on RGB images, their assumptions and modeling strategies do not generalize well to hyperspectral images~(HSI), which are inherently high-dimensional and exhibit complex spectral structures. In this paper, we propose the first GCD framework tailored for HSI, introducing a prototype subspace modeling model to better capture class structure. Instead of learning a single prototype vector for each category as in existing methods such as SimGCD, we model each category using a set of basis vectors, forming a subspace representation that enables greater expressiveness and discrimination in a high-dimensional feature space. To guide the learning of such bases, we enforce two key constraints: (1) a basis orthogonality constraint that promotes inter-class separability, 
and (2) a reconstruction constraint that ensures each prototype basis can effectively reconstruct its corresponding class samples. Experimental results on real-world HSI
demonstrate that our method significantly outperforms state-of-the-art GCD methods, establishing a strong foundation for generalized category discovery in hyperspectral settings.
\end{abstract}


\section{Introduction}
Hyperspectral images (HSI) capture detailed spectral information across hundreds of contiguous wavelength bands, enabling fine-grained material discrimination beyond the capabilities of conventional RGB imagery. This unique characteristic has led to widespread adoption of HSI in various domains, including land cover classification, environmental monitoring, and precision agriculture~\cite{khan2018modern}.

In hyperspectral image classification, both supervised and unsupervised learning approaches have been widely explored. Supervised methods, which depend on labeled samples, often achieve high accuracy~\cite{10101853, 9448337}; however, obtaining reliable annotations for hyperspectral data is costly and labor-intensive, limiting their scalability in real-world scenarios. In contrast, unsupervised methods~\cite{9191074, huang2023model} cluster data based on intrinsic structures without requiring labels, but they often lack semantic guidance and struggle to differentiate spectrally similar classes, particularly in the presence of noise. To address these limitations, generalized category discovery (GCD) has emerged as a compelling alternative~\cite{Vaze_2022_CVPR}. Unlike conventional approaches that assume a fixed set of categories—either all known (supervised) or all unknown (unsupervised)—GCD adopts a more realistic and flexible assumption: the unlabeled data may comprise both known and novel categories. The objective is to correctly recognize samples from known classes while simultaneously discovering new, unseen categories. This category-partial setting aligns well with real-world hyperspectral scenarios, especially in static remote sensing applications where the number and composition of observed materials can vary across scenes, and labeled categories only partially represent the actual distribution.


While GCD has been extensively studied in the context of RGB image classification~\cite{Vaze_2022_CVPR, wen2023simgcd, pu2023dynamic, zhang2023promptcal}, it remains largely unexplored in hyperspectral image analysis, where the data exhibits fundamentally different characteristics. Hyperspectral images typically consist of hundreds of spectral bands, leading to high dimensionality and fine-grained spectral variation. These challenges are further compounded by the lack of large-scale annotated datasets, which prevents the use of pretrained backbones commonly adopted in RGB-based GCD methods. As a result, directly applying such methods to hyperspectral data is ineffective, especially under limited supervision and in the absence of strong semantic priors. For example, SimGCD~\cite{wen2023simgcd} represents each class using a single prototype vector, which has proven effective on RGB image datasets by capturing the semantic center of each category. However, this assumption is inadequate for hyperspectral data, where a single vector is insufficient to capture the complex and high-dimensional structure of each class, particularly in the absence of powerful feature extractors or dimensionality reduction backbones. 

As the first work addressing hyperspectral generalized class discovery (GCD), we propose a novel approach that learns category-specific prototype bases rather than single prototype vectors. Each basis spans a low-dimensional subspace that represents the intrinsic structure of a class in the high-dimensional spectral space. To ensure that these learned bases are both expressive and discriminative, we impose two key constraints: (1) an orthogonality constraint that promotes inter-class separability, and (2) a reconstruction constraint that encourages each basis to accurately represent its corresponding samples. This structure-aware design is well-suited for hyperspectral data and shows promising results on benchmark datasets, surpassing several adapted GCD methods. Furthermore, our method inherits the linear complexity of SimGCD~\cite{wen2023simgcd}, while introducing a basis representation module that incurs only a constant-factor increase in computation in latent space, thereby preserving scalability to large-scale unlabeled datasets.

\section{Related Work}

In this section, we review related work on generalized category discovery and subspace modeling that informs our method for hyperspectral image analysis.

\subsection{Generalized Category Discovery~(GCD)}

Consider a dataset consisting of a labeled subset \( \mathcal{D}_l = \{(x_i^l, y_i^l)\} \), where \( x_i^l \in \mathbb{R}^d \) denotes an input feature and \( y_i^l \in \mathcal{Y}_l \) is its corresponding label from a known class set \( \mathcal{Y}_l \), and an unlabeled subset \( \mathcal{D}_u = \{x_j^u\} \) that contains samples from both known classes \( \mathcal{Y}_l \) and unknown (novel) classes \( \mathcal{Y}_u \), with \( \mathcal{Y}_l \cap \mathcal{Y}_u = \emptyset \). 
The goal of GCD is to learn a representation space that is discriminative for both known classes and unknown ones. Specifically, the model should (1) align samples in \( \mathcal{D}_u \) belonging to known classes with their corresponding categories in \( \mathcal{Y}_l \), and (2) partition the remaining samples into distinct, coherent clusters, each ideally corresponding to a novel class in \( \mathcal{Y}_u \), despite the absence of ground-truth labels, where each discovered cluster is assigned an arbitrary index and may later be interpreted through semantic analysis or manual inspection.

As the first work to formalize the GCD setting, Vaze et al.~\cite{Vaze_2022_CVPR} proposed a non-parametric framework that learns features from both labeled and unlabeled samples with a neural network, and then applies a semi-supervised K-means algorithm to assign instances to known and novel categories. SimGCD~\cite{wen2023simgcd} streamlines GCD by replacing the semi-supervised K-means with a parametric network, which learns a set of prototype vectors—each representing a category—and assigns instances based on their similarity to these prototypes. Based on SimGCD, a dual-context framework~\cite{luo2024contextuality} incorporated both instance- and cluster-level contextual information into representation learning, enhancing feature quality through structured context modeling. PromptGCD \cite{zhang2023promptcal} augments the Vision Transformer with additional trainable inputs that help distinguish different semantic categories. DCCL~\cite{pu2023dynamic} proposes a hierarchical contrastive learning framework for GCD, where class-level prototypes are built from labeled data and class prototypes are dynamically discovered from unlabeled data using graph-based clustering approach.

Despite these advances, GCD has not been systematically explored in the hyperspectral images domain, where data is high-dimensional and exhibits rich spectral-spatial structures. 
Our work fills this gap by proposing a GCD framework tailored for hyperspectral image analysis.

\subsection{Subspace Modeling for Hyperspectral Images}

Subspace modeling assumes that high-dimensional data lies in a union of low-dimensional subspaces, where each subspace corresponds to a distinct semantic class~\cite{elhamifar2013sparse}. It has achieved remarkable success in high-dimensional data analysis, particularly in the context of HSI~\cite{huang2023model}. Two major approaches to hyperspectral subspace modeling are self-representation and basis-representation models.

Self-representation models express each data point as a linear combination of other data points, typically under sparse or low-rank constraints~\cite{huang2018joint, zhai2018laplacian}. The resulting representation coefficients form a pairwise affinity matrix that reflects the similarities or dependencies between data points, making them well-suited for building graph-based clustering models. However, this formulation often incurs high computational complexity due to the size of the coefficient matrix~\cite{huang2020sketch}.

In contrast, basis-representation models learn a compact set of shared bases and represent each data point as a linear combination of these bases~\cite{li2023model}. The coefficients here do not reflect pairwise relations between data points, but instead capture how each point is associated with the underlying subspaces. This structure avoids computing pairwise affinities, significantly reducing complexity and enhancing scalability, especially for large-scale hyperspectral datasets and end-to-end training.

\section{METHODOLOGY}
\begin{figure}[ht]
    \centering
    \includegraphics[width=0.8\linewidth]{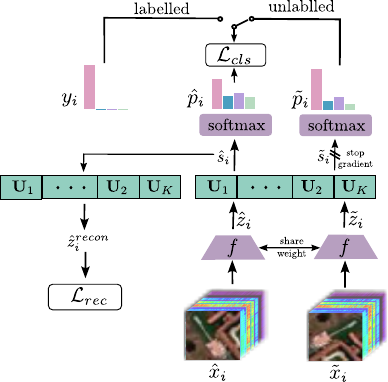}
    \caption{Structure of the proposed method. Two augmented views, $\hat{x}_i$ and $\tilde{x}_i$, are encoded by a shared-weight autoencoder and projected onto shared prototype bases $\mathbf{U}$. The projection coefficients are used for classification, supervised by either ground-truth labels or stop-gradient pseudo-labels from a low-temperature softmax. These coefficients also guide the reconstruction of input features. The model is optimized with classification loss $\mathcal{L}_{cls}$ and reconstruction loss $\mathcal{L}_{rec}$.}
    \label{fig:structure}
\end{figure}

In this section, we first introduce a strong baseline for GCD in natural RGB images, where each class is represented by a single prototype vector. Building on this baseline, we extend the model to hyperspectral image data by representing each class as a subspace spanned by a set of basis vectors. This modification enables a more expressive and flexible representation, better capturing the complex spectral variations inherent to hyperspectral images. The overall structure is illustrated in Fig.~\ref{fig:structure}.

\subsection{Preliminary Model on RGB Images: SimGCD}
SimGCD~\cite{wen2023simgcd} is a representative prototype-based GCD framework that achieves strong performance on RGB image benchmarks. It represents each known class with a learnable prototype vector and applies contrastive learning and entropy regularization to optimize jointly representation learning and classification networks. 
This model is composed mainly of two parts: (1) a representation learning network and (2) a classifier learning network.

\textbf{Representation Learning Network.}
The goal of the representation learning network is to extract discriminative features that facilitate accurate classification. 
In the original SimGCD framework, a pretrained Vision Transformer (ViT) from DINO~\cite{caron2021emerging} is adopted as the backbone and finetuned via contrastive learning using both labeled and unlabeled data.

Specifically, for each input image $x_i$, two randomly augmented views $\hat{x}_i$ and $\Tilde{x}_i$ are generated. 
These views are independently passed through the encoder $f(\cdot)$ to obtain features:
\begin{equation}
    \hat{z}_i = f(\hat{x}_i), \quad \Tilde{z}_i = f(\Tilde{x}_i). 
\end{equation}
A contrastive loss is then applied to bring the two augmented views $\hat{z}_i$ and $\Tilde{z}_i$ of the same instance closer in the feature space, while pushing apart representations of different instances.  Formally, the unsupervised contrastive loss follows a SimCLR-style~\cite{chen2020simple} formulation:
\begin{equation}
    \mathcal{L}_{rep}^{u} = \frac{1}{B} \sum\limits_{\hat{z}_j \in B}  -\log \frac{\exp(\mathrm{sim}(\hat{z}_i, \Tilde{z}_i)/\tau_u)}{\sum\limits_{\hat{z}_j \in B} \exp(\mathrm{sim}(\hat{z}_j, \Tilde{z}_i)/\tau_u)}
\end{equation}
where $\mathrm{sim}(\mathbf{x}, \mathbf{y}) = \frac{\mathbf{x}^\top \mathbf{y}}{\|\mathbf{x}\| \, \|\mathbf{y}\|}$ denotes cosine similarity, and $\tau_u$ is a temperature parameter. To further leverage the supervision from labeled samples, a supervised contrastive loss~\cite{khosla2020supervised} is employed. The supervised contrastive loss is defined as:
\begin{equation}
    \mathcal{L}_{rep}^{s} = \sum_{i \in B_{l}} \frac{1}{|\mathcal{P}_i|} \sum_{\Tilde{z}_j \in \mathcal{P}_i} -\log \frac{\exp(\mathrm{sim}(\hat{z}_i, \Tilde{z}_j)/\tau_s)}{\sum\limits_{n\neq i} \exp(\mathrm{sim}(\hat{z}_i, \Tilde{z}_n)/\tau_s)},
\end{equation}
where $B_{l}$ denotes the set of labeled data, $\tau_s$ is the temperature parameter, $\mathcal{P}_i$ is the positive set (data share same label with $z_i$). This dual contrastive training strategy enables the model to effectively leverage both labeled and unlabeled data. And the loss function of representation learning is defined as below:
\begin{equation}
    \mathcal{L}_{\text{rep}} = (1 - \lambda)\mathcal{L}_{\text{rep}}^{u} + \lambda \mathcal{L}_{\text{rep}}^{s}
\end{equation}
where $\lambda$ is the weight to balance the effect of different parts.

\textbf{Classifier Learning Network.}
In SimGCD, a parametric classifier is trained to assign samples to their corresponding categories based on their similarity to class prototypes. Formally, given a dataset $\mathcal{Y}_l \cup \mathcal{Y}_u$ with $K$ distinct classes, a set of $K$ prototype vectors $\mathbf{C} = [\mathbf{C}_1, \mathbf{C}_2, \cdots, \mathbf{C}_K]$ is randomly initialized. For a latent feature $\hat{z}_i$, the soft label is computed based on the cosine similarity between the feature and the prototype vectors as follows: 
\begin{equation}
\hat{p}_i^{(k)} = \frac{\exp(\mathrm{sim}(\hat{z}_i, \mathbf{C}_k)/{\tau_c})}{\sum_{k} \exp(\mathrm{sim}(\hat{z}_i, \mathbf{C}_{k})/\tau_c)},
\end{equation}
where $\tau_c$ is a temperature parameter. The soft pseudo-label $\Tilde{p}_i^{(k)}$ is computed using a smaller temperature. The classification objectives include the cross-entropy loss between predictions and either soft pseudo-labels or ground-truth labels:
\begin{equation}
\mathcal{L}_{cls}^{u} = \frac{1}{|B|}\sum_{i \in |B|} \mathcal{L}_{ce}(\hat{p}_i, \Tilde{p}_i) - \epsilon H(\overline{p}),
\end{equation} 
\begin{equation}
\mathcal{L}_{cls}^{s} = \frac{1}{|B|_l} \sum_{i \in |B|_l} \mathcal{L}_{ce}(y_i, \hat{p}_i), 
\end{equation}
where $\mathcal{L}_{ce}$ denotes the cross-entropy loss function. The term $\overline{p} = \frac{1}{2B}\sum{i\in B}(\hat{p}_i + \Tilde{p}_i)$ represents the mean prediction of a batch, and $H(\overline{p}) = -\sum_k \overline{p}^{(k)} \log(\overline{p}^{(k)})$ is the entropy of the batch distribution, which encourages balanced cluster assignments and prevents collapse into a small number of dominant clusters. To integrate both supervised and unsupervised signals, the classification loss is defined as:
\begin{equation}
    \mathcal{L}_{\text{cls}} = (1 - \lambda)\mathcal{L}_{\text{cls}}^{u} + \lambda \mathcal{L}_{\text{cls}}^{s}.
\end{equation}
Here, \(\mathcal{L}_{\text{cls}}^s\) and \(\mathcal{L}_{\text{cls}}^u\) denote the classification losses for labeled and unlabeled data, respectively, and \(\lambda \in [0,1]\) balances their contributions. In our model, $\lambda$ is empirically set to 0.4, and can be fine-tuned for different datasets or tasks.

\subsection{Prototype Subspace Modeling}
Existing generalized class discovery (GCD) approaches for RGB imagery predominantly rely on feature‐extraction backbones pretrained on large‐scale datasets (e.g., ImageNet), which provide robust initialization for downstream tasks. By contrast, hyperspectral images, which comprise hundreds of spectral bands, lack such pretrained backbones due to the scarcity of annotated hyperspectral data, making it infeasible to directly apply RGB-based methods to hyperspectral GCD.

To address these issues, we are motivated to develop a subspace modeling approach that better captures the spectral structure of hyperspectral data and enables effective learning even in the absence of pre-trained weights.

\textbf{Basis Representation Model.} High-dimensional data is rarely uniformly distributed in the ambient space. Instead, it tends to concentrate along low-dimensional subspaces. In many cases, data from a particular category can be effectively represented by a low-dimensional subspace of the high-dimensional space~\cite{elhamifar2013sparse}. Specifically, let $\mathbf{U} = [ \mathbf{U}_1, \mathbf{U}_2, \ldots, \mathbf{U}_K]$,
where each $\mathbf{U}_k \in \mathbb{R}^{d \times r}$ denotes the basis prototype for category $\mathcal{C}_k$, with $r \ll d$ representing the subspace and data dimensions, respectively.

To encourage well-separated subspaces and reduce overlap between categories, the basis prototypes $\mathbf{U}_{k}$ are required to be pairwise orthogonal:
\begin{equation}
\mathbf{U}_k^\top \mathbf{U}_m = \mathbf{0}, \quad \forall k \ne m.
\end{equation}
When $k = m$, the self-correlation $\mathbf{U}_k^\top \mathbf{U}_k$ is not penalized, as it reflects the internal structure of each subspace and does not affect inter-subspace separation.
Furthermore, each \(\mathbf{U}_k\) serves as a representative basis for samples belonging to category \(\mathcal{C}_k\). Given a high-dimensional latent feature \(\hat{z}_i \in \mathbb{R}^{1\times d}\), if the sample belongs to category \(\mathcal{C}_k\), its representation can be approximated using \(\mathbf{U}_k\) as
\begin{equation}
\hat{z}_i^\top \approx \mathbf{U}_k \mathbf{V}_i,
\end{equation}
where \(\mathbf{V}_i\) denotes the representation coefficients in the subspace spanned by \(\mathbf{U}_k\).

Intuitively, if $\hat{z}_i$ can be well represented by a subspace, it will exhibit stronger alignment with its basis. This alignment can be quantified by projecting \(z_i\) onto each \(\mathbf{U}_k\), where a larger projection indicates a stronger correlation with the subspace, and thus a higher likelihood of category membership~\cite{cai2022efficient}.
To infer the probability of \(\hat{z}_i\) belonging to each subspace, we measure its alignment with all basis prototypes by computing the projection onto each subspace. Specifically, we compute the projection score as
\begin{equation}
s_i^k = \| \hat{z}_i\mathbf{U}_k \|_2^2,
\end{equation}
which quantifies how well \(\hat{z}_i\) aligns with subspace \(\mathcal{C}_k\). The 
$p_i$ is thereby redefined as a soft label via softmax normalization, as follows:
\begin{equation}
p_i^k = \frac{\exp(s_i^k/\tau_c)}{\sum_{m=1}^K \exp(s_i^m/\tau_c)},
\end{equation}
where \(p_i^k\) denotes the estimated probability that \(\hat{z}_i\) belongs to category \(\mathcal{C}_k\). This subspace-based prototype representation enhances the discriminative capacity of the latent space and improves category discovery by leveraging the intrinsic structure of hyperspectral data, which naturally reside on multiple low-dimensional manifolds associated with distinct material categories.

\textbf{Optimization Objectives.}
To ensure that the learned subspace bases are pairwise orthogonal, we adopt an orthogonality loss similar to~\cite{cai2022efficient}, defined as:
\begin{equation} 
\label{constrain} 
\mathcal{L}_{\text{orth}} = \left\| (\mathbf{U}^\top \mathbf{U}) \odot \mathbf{O} \right\|_F^2,
\end{equation}
where $\odot$ denotes the Hadamard (element-wise) product, and \mbox{$\|\cdot\|_F$} denotes the Frobenius norm. The binary mask matrix $\mathbf{O} \in \{0,1\}^{Kr \times Kr}$ is used to exclude intra-subspace inner products and only penalize inter-subspace correlations.

During training, the SimGCD framework encourages the soft-label $p_i$ to become increasingly sharp, often approaching a state where one basis dominates the representation. To simplify computation, we approximate the reconstruction process by projecting each feature onto the entire basis set:
\begin{equation}
    \mathcal{L}_{\text{rec}} =\frac{1}{B} \sum_{i \in |B|} \left\| z_i - \mathbf{U} \mathbf{U}^\top z_i \right\|_2^2,
\end{equation}
Here, $\mathbf{U}\mathbf{U}^{\top}$ is a projection matrix: it projects $z_i$ onto the union of all category subspaces and then reconstructs it within that span, so the loss simply measures the distance between each feature and its basis reconstruction value. Although the projection uses the full basis, the self-distillation process typically sharpens the assignment distribution, leading to reconstructions that effectively align with a single subspace.

\subsection{Overall Objective}
We integrate the proposed subspace modeling objectives with supervised and unsupervised learning components. The full loss is:
\begin{equation}
    \mathcal{L}_{all} = \mathcal{L}_{rep} + \mathcal{L}_{cls} + \mathcal{L}_{\text{orth}} + \mathcal{L}_{\text{rec}}.
\end{equation}
Here, \( \mathcal{L}_{\text{rep}} \) denotes the representation learning loss, and \( \mathcal{L}_{\text{cls}} \) is the classification loss. \( \mathcal{L}_{\text{orth}} \) enforces orthogonality among subspace bases, while \( \mathcal{L}_{\text{rec}} \) ensures that the subspace bases can faithfully reconstruct the input. In practice, we empirically set all loss weights to 1. This setting works well across datasets and avoids introducing additional hyperparameter tuning, though the weights can be adjusted if needed.

\section{Experiments}

In this section, we present the experimental setup and results to evaluate the effectiveness of our proposed method.

\subsection{Setting}
\textbf{Datasets Setting.} We evaluate our model on 3 benchmark datasets as shown in Table~\ref{tab:dataset_summary}

\begin{table}[ht]
\centering
\caption{Summary of hyperspectral datasets and experimental settings}
\label{tab:dataset_summary}
\resizebox{\columnwidth}{!}{%
\begin{tabular}{@{}lccc@{}}
\toprule
\textbf{Attribute}          & \textbf{Salinas}                & \textbf{Trento}                  & \textbf{PaviaU}                     \\
\midrule
Image size                  & $512 \times 217$           & $600 \times 166$        & $610 \times 340$             \\
Spectral bands              & 204              & 63                                & 103                  \\
Number of classes           & 16             & 6              & 9                         \\
Labeled classes             & [0--8]                            & [0--3]                            & [0--4]               \\
Unlabeled classes           & [8--16]                           & [3--6]                            & [4--9]                    \\
Label ratio                 & 0.5                               & 0.5                               & 0.5                          \\
Train samples             & 43303                           & 24171                      & 34220                          \\
Test samples             & 10826                            & 6043                     & 8556                            \\
\bottomrule
\end{tabular}%
}
\end{table}
As shown in the table, we consider a setting similar to previous works on RGB datasets which provide half of classes label information and then sample 50\% of the data points as $\mathcal{D}_l$. 

\subsection{Backbone and Data Augmentation}
In the absence of pre-trained feature extractors for hyperspectral imagery (HSI), we employ a lightweight CNN-based encoder to efficiently capture spectral–spatial features. Compared with deeper architectures such as ViT or ResNet, lightweight CNNs are less susceptible to overfitting on small patch-based inputs and can be trained stably from scratch, while also reducing computational overhead and maintaining sufficient modeling capacity for patch-level clustering, where non-local context is inherently limited. To prepare the inputs, for each pixel in the HSI we extract a patch centered on the target pixel and generate two augmented views, which are used as paired inputs to our model. To ensure the spatial alignment of the center pixel, we avoid augmentations that alter its position. Consequently, only spatially invariant augmentations such as random noise, rotation, and flipping are applied. More details can be found in our code repository\footnote{https://github.com/lxlscut/gcd\_hsi.}.


\subsection{Experiment Results}
In this section, we compare our method with several competing approaches: 
K-means~\cite{macqueen1967some}, which clusters the features learned from both labeled and unlabeled data; 
GCD~\cite{Vaze_2022_CVPR}, employing contrastive learning for generalized category discovery; 
SimGCD~\cite{wen2023simgcd}, a parametric classification method with entropy regularization; 
Contextuality-GCD~\cite{luo2024contextuality}, leveraging instance and cluster-level contexts for enhanced representation learning; 
and DCCL~\cite{pu2023dynamic}, implementing dynamic conceptual contrastive learning for clustering unlabeled data. 
To all methods, we apply the same feature extraction network and data augmentation strategy. 
Results are shown in \cref{tab:xp1}.
\begin{table}[ht]
\centering
\caption{Results of different methods on evaluation datasets.}
\label{tab:xp1}
\resizebox{0.48\textwidth}{!}{
\begin{tabular}{l|ccc|ccc|ccc}
\hline
\textbf{Methods} & \multicolumn{3}{c|}{\textbf{Trento}} & \multicolumn{3}{c|}{\textbf{Salinas}} & \multicolumn{3}{c}{\textbf{PaviaU}} \\
                 & All & Old & New & All & Old & New & All & Old & New \\
\hline
K-means~\cite{macqueen1967some}    & 60.70 & 91.62 & 50.67 & 73.92 & \textbf{99.95} & 40.06 & 60.62 & 57.64 & 68.09 \\
GCD~\cite{Vaze_2022_CVPR} & 64.16 & \textbf{100.00} & 52.53 & 73.33 & 97.27 & 42.19 & 58.51 & 71.16 & 26.88 \\
DCCL~\cite{pu2023dynamic} & 79.53 & 99.80 & 72.96 & 77.30 & 74.97 & \textbf{80.32} & 59.47 & 59.47 & 59.45 \\
SimGCD~\cite{wen2023simgcd}  & 82.87 & 78.24 & 84.37 & 69.14 & 76.28 & 59.85 & 51.75 & 46.14 & 65.79 \\
Contextual-GCD~\cite{luo2024contextuality} & 94.11 & 98.51 & 92.68 & 64.47 & 72.32 & 54.26 & 58.68 & 60.23 & 54.83 \\
Ours   & \textbf{94.87} & 97.70 & \textbf{93.95} & \textbf{83.30} & 93.60 & 69.90 & \textbf{79.51} & \textbf{82.46} & \textbf{72.14} \\
\hline
\end{tabular}
}
\captionsetup{justification=raggedright, singlelinecheck=false}
\caption*{\footnotesize \textbf{Note:} "Old" = known classes, "New" = novel classes, "All" = overall accuracy. Bold indicates the best result.}
\end{table}

From the experimental results, we observe that applying K-means clustering on the learned feature representations can achieve reasonable performance, benefiting from the initial feature extraction. GCD improves upon this by combining feature learning with a semi-supervised K-means step, using partial label information to guide clustering. This enhances the alignment with old classes and improves performance on datasets like Trento and PaviaU, although the rigid nature of K-means limits its flexibility on more diverse datasets such as Salinas.
SimGCD further advances the design by replacing the semi-supervised K-means step with a parameterized prototype-based classifier, enabling end-to-end optimization. Nevertheless, modeling each class with a single prototype vector proves insufficient for capturing the data present in high-dimensional space, leading to weaker overall performance. Contextual-GCD enhances feature learning by incorporating dual-context contrastive clustering, leveraging both instance-level and prototype-level contexts to improve category discovery in mixed-class unlabeled datasets. While this enhances feature discrimination and robustness, leading to improved performance on Trento and PaviaU, the added regularization may impair the model’s capacity to handle complex distributions, resulting in a slight decline in performance on the Salinas dataset. DCCL achieves better results than SimGCD, but it requires additional computational resources due to the overhead introduced by dynamic conception generation and graph-based clustering, making it less efficient for large-scale datasets. Through subspace modeling of prototypes, our method directly captures the underlying structure of hyperspectral data by learning a compact and discriminative subspace basis for different classes. Benefiting from this modeling, our method achieves superior performance on overall accuracy, consistently outperforming existing baselines on Trento, Salinas, and PaviaU datasets with strong balance between old and novel category discovery.

\subsection{Ablation Study}
In this section, we compare the baseline model SimGCD with our proposed method under different constraint configurations. The results are presented in \cref{tab:ssb}.
\begin{table}[!h]
\centering
\caption{Results of ablation study}
\label{tab:ssb}
\resizebox{0.5\textwidth}{!}{
\begin{tabular}{l|ccc|ccc|ccc}
\hline
\textbf{Methods} & \multicolumn{3}{c|}{\textbf{Trento}} & \multicolumn{3}{c|}{\textbf{Salinas}} & \multicolumn{3}{c}{\textbf{PaviaU}} \\
                 & All & Old & New & All & Old & New & All & Old & New \\
\hline
SimGCD~\cite{wen2023simgcd}      & 82.87 & 78.24 & 84.37 & 69.14 & 76.28 & 59.85 & 51.75 & 46.14 & 65.79 \\
\hline
Ours w/o $\mathcal{L}_o$\&$\mathcal{L}_{r}$    & 70.23 & 66.01 & 71.60 & 66.31 & 74.74 & 55.35 & 53.41 & 47.53 & 68.13 \\
Ours w/o $\mathcal{L}_o$       & 86.13 & 69.53 & 91.52 & 68.20  & 75.02 & 59.32 & 57.35 & 51.05 & \textbf{73.12} \\
Ours w/o $\mathcal{L}_{r}$       & 79.99 & 64.66 & 84.97 & 65.27 & 73.73 & 54.26 & 52.86 & 47.41 & 66.49 \\
\hline
Ours     & \textbf{94.87} & \textbf{97.70} & \textbf{93.95} & \textbf{83.30} & \textbf{93.60} & \textbf{69.90} & \textbf{79.51} & \textbf{82.46} & 72.14 \\
\hline
\end{tabular}
}
\captionsetup{justification=raggedright, singlelinecheck=false}
\caption*{\footnotesize \textbf{Note:} "Old" = known classes, "New" = novel classes, "All" = overall accuracy. Bold indicates the best result.}
\end{table}

We observe that simply replacing the change vector with a set of vectors does not effectively capture the underlying data structure, as shown by the performance drop in the our method without the orthogonality and reconstruction losses. Moreover, we find that both $\mathcal{L}_o$ and $\mathcal{L}_{r}$ constraints play complementary roles during training. When either constraint is removed, the performance deteriorates across all datasets, indicating that relying on only one of them is insufficient. In contrast, applying both constraints jointly yields significant performance improvements and achieves the best overall results.

\begin{figure}[h]
  \centering
  \begin{tikzpicture}
    \begin{axis}[
      width=0.9\columnwidth,
      height=5.2cm,
      xlabel={$\epsilon$ Value},
      ylabel={ACC (\%)},
      xmin=0, xmax=100,
      ymin=30, ymax=100,
      xtick={0,20,40,60,80,100},
      ytick={30,40,50,60,70,80,90,100},
      legend pos=south east,
      grid=major,
      grid style={dashed,gray!30},
      line width=1.5pt,
      mark size=3pt,
      axis lines=left,
      axis line style={thick,black},
      tick style={thick,black},
      tick label style={font=\small},
      label style={font=\normalsize},
    ]

    \addplot[
      color=blue!80,
      mark=*,
      mark options={fill=blue!80},
      smooth
    ] coordinates {
      (0,47.29)   (20,89.81)
      (40,88.70)  (60,94.87)
      (80,97.07)  (100,88.15)
    };
    \addlegendentry{\textbf{Trento}}

    \addplot[
      color=red!80,
      mark=square*,
      mark options={fill=red!80},
      smooth
    ] coordinates {
      (0,74.63)   (20,63.92)
      (40,83.64)  (60,79.51)
      (80,60.04)  (100,57.30)
    };
    \addlegendentry{\textbf{PaviaU}}



    \end{axis}
  \end{tikzpicture}
  \caption{Effect of the regularization parameter $\epsilon$}
  \label{fig:epsilon_performance_enhanced}
\end{figure}
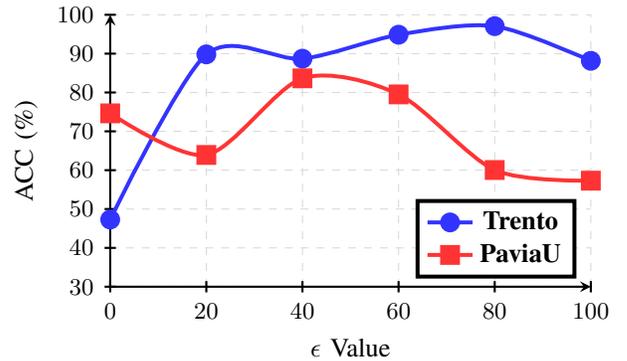

\begin{figure*}[t]
    \centering
    \includegraphics[width=0.9\linewidth]{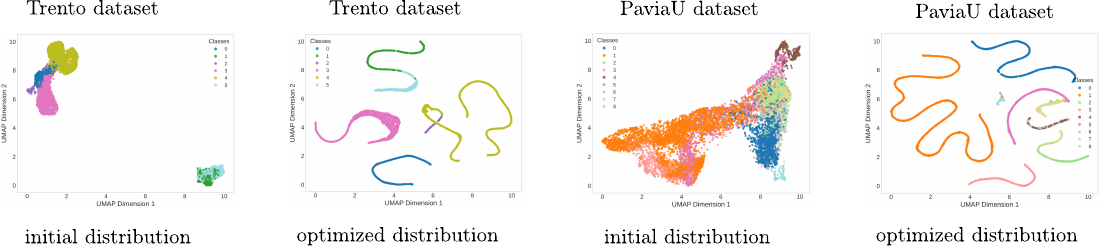}
    \caption{Cluster distributions on Trento and PaviaU datasets}
    \label{fig:distribution}
\end{figure*}

\subsection{Parameter Study}
The parameter $\epsilon$ controls the strength of entropy-based regularization to mitigate prediction bias by encouraging appropriate class distribution during learning. We evaluated its effect on the Trento and PaviaU dataset, as shown in Fig.~\ref{fig:epsilon_performance_enhanced}. Results show that both very small and very large values of $\epsilon$ degrade clustering performance. A large $\epsilon$ leads to overly uniform predictions, while a small one can miss small clusters—though it may sometimes boost accuracy. In our experiments, we fixed $\epsilon = 60$ across all datasets to ensure accurate cluster detection.

\subsection{Visual Results}
To provide a clearer view of the clustering evolution, Fig.~\ref{fig:distribution} visualizes the cluster distributions on the Trento and PaviaU datasets. In the initial stage (left panels), clusters are heavily mixed and spatially entangled, making boundaries indistinct. After optimization (right panels), the separation improves substantially, with most clusters forming compact, well-defined regions and only a few exhibiting slight overlap. This clear transition underscores the effectiveness of the learning process.

\section{Conclusion}
In this work, we present the first generalized category discovery (GCD) framework specifically designed for hyperspectral images. Specifically, we propose a prototype basis learning model that models each category as a subspace spanned by a few basis vectors, rather than a single prototype point as in conventional methods. To guide the learning of such structured prototypes, we introduce two complementary constraints: a basis orthogonality constraint to promote inter-class separability, and a reconstruction constraint to ensure the representation ability of each class basis. Ablation studies demonstrate that these constraints effectively supervise the learning of meaningful and discriminative bases. Extensive experiments on multiple HSI benchmarks validate the effectiveness of our approach, consistently outperforming state-of-the-art GCD methods and demonstrating the advantages of structured prototype modeling in capturing the spectral structure and class distinctions in hyperspectral data. 
\bibliographystyle{IEEEtran}
\bibliography{reference}
\end{document}